\documentclass[letterpaper, 10 pt, conference]{ieeeconf}  

\let\proof\relax

\let\tempindent\labelindent
\let\labelindent\relax

\usepackage{amsthm}
\usepackage{amssymb,amsmath,amsfonts}
\usepackage{Definitions}
\usepackage{booktabs}
\usepackage{multirow}
\usepackage{makecell}
\usepackage{graphicx}
\usepackage{subfigure}
\usepackage{wrapfig}
\usepackage{float}
\usepackage{bm}
\usepackage{multicol}

\usepackage{enumitem}
\let\labelindent\tempindent
\let\tempindent\relax

\IEEEoverridecommandlockouts                              

\title{\LARGE \bf
A Single Diffusion-Policy Controller for\\
Multi-Task Block Pushing with Zero-Shot Sim-to-Real Transfer
}

\author{
Haitong Ma$^{1}$, Haldun Balim$^{1}$, Yang Hu$^{1}$, Bo Dai$^{2}$ and Na Li$^{1}$
\thanks{$^{1}$Harvard University.}
\thanks{$^{2}$Georgia Institute of Technology.}
}

\usepackage{color,xcolor}

\allowdisplaybreaks[4]

\begin{document}

\maketitle
\thispagestyle{empty}
\pagestyle{empty}

\begin{abstract}
Diffusion policies have shown promising empirical performance in representing and learning complex maneuvers for robots using behavior cloning (BC).
In this paper, we explore training diffusion policies from scratch using reinforcement learning (RL) for multi-task robotic manipulation.
Specifically, we aim to train \emph{a single diffusion policy} for block-pushing tasks with multiple shapes. 
The proposed framework features a simple policy loss function, which is a reweighted evidence lower bound used in BC-based diffusion policy training and can seamlessly serve as the policy learning module in RL algorithms. 
To address the exploration challenges arising from the absence of demonstrations, we incorporate reverse curriculum generation and objective-centric representations. Combined with the expressiveness of diffusion policies, our design supports learning of multi-task block-pushing policies in our sparse-reward simulation setting.
We further evaluate whether the trained diffusion policy transfers \emph{in zero-shot} to real-world tasks under varying environmental conditions including goal positions, block shapes, block weights and surface friction, providing evidence that this pipeline can transfer to our real-world block-pushing setup under the tested variations.
\end{abstract}



\section{Introduction}
Denoising diffusion probabilistic model (DDPM)~\cite{ho2020denoising,sohl-dickstein2015deep,song2021scorebased} has been showing remarkable expressiveness and flexibility in representing and generating complex data distributions~\cite{dhariwal2021diffusion}. In light of these strengths, diffusion policies have been widely leveraged to generate long-horizon robot trajectories~\cite{janner2022planning,chi2023diffusion,du2024learning} or imitate expert policies in MDPs~\cite{wang2022diffusion,chen2022offline,hansen2023idql,zhang2025energy} using behavior cloning or offline RL methods when expert demonstrations are available in abundance.

Building on this foundation, a natural question to ask is whether such expressive policy classes can go beyond single-task learning and serve as a basis for more general-purpose robotic intelligence, since the ultimate goal of autonomous robotic manipulation is to learn one single universal policy that can solve a variety of different tasks~\cite{du2024learning}. These tasks are often \emph{contact-rich}, where the presence of frequent and complex contacts makes the planning landscape highly sensitive to differences in both task objectives and external disturbances. Capturing such variability requires sufficiently expressive models to parameterize the universal policies—this is precisely where diffusion policies come into play, owing to their inherent expressiveness and flexibility~\cite{he2023diffusion,ni2023metadiffuser}.



Despite the great success of diffusion policies, the current training approaches are generally based on behavior cloning (BC)~\cite{chi2023diffusion}, while RL only serves as an optional fine-tuning method~\cite{janner2022planning}. However, such BC-based approaches could be challenging for multi-task manipulations, since the task space requires a lot of diverse expert demonstrations across different tasks, which limits the scalability of the BC approaches. Worse still, the fine-tuning phase is usually fragile and faces the challenge of catastrophic forgetting~\cite{kirkpatrick2017overcoming}, hindering its ability to reliably adapt to new tasks without compromising the performance of previously learned skills.

Recently, a growing interest has emerged in training diffusion policies from scratch using online RL methods~\cite{yang2023policy,ding2024diffusion,ren2024diffusion}, which achieves stronger performance compared to previous algorithms based on deterministic or Gaussian policies~\cite{schulman2017proximal,haarnoja2018soft}. In fact, online RL has shown its strong potential for multi-task learning in arcade games and robotics locomotion~\cite{espeholt2018impala,margolis2023walk}.
However, multi-task diffusion has been studied only with  BC-based training~\cite{he2023diffusion,ma2024hierarchical} but not yet with online RL. Meanwhile, one of the major challenges of applying online RL to real-world settings lies in sim-to-real transfer~\cite{tobin2017domain}. To the best of our knowledge, few studies have studies sim-to-real transfer for diffusion policies.
%


\begin{figure*}[t]
    \centering
    \includegraphics[width=0.75\linewidth]{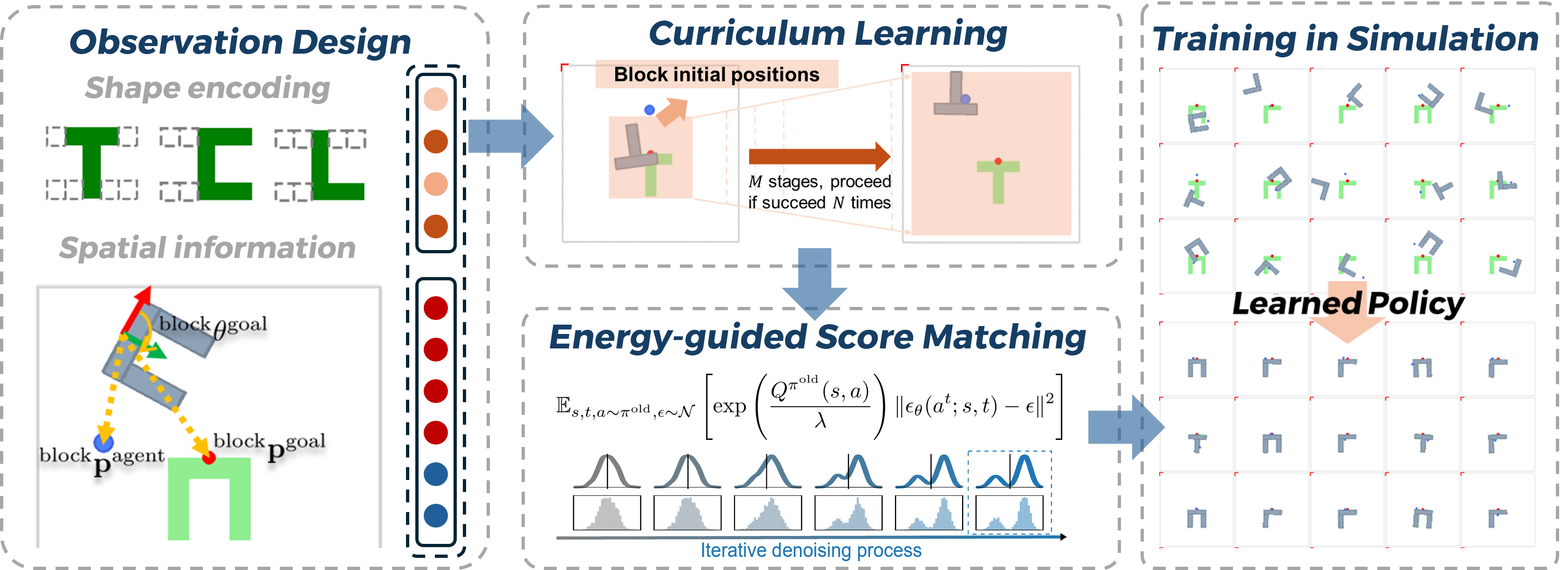}\\
    \includegraphics[width=0.75\linewidth]{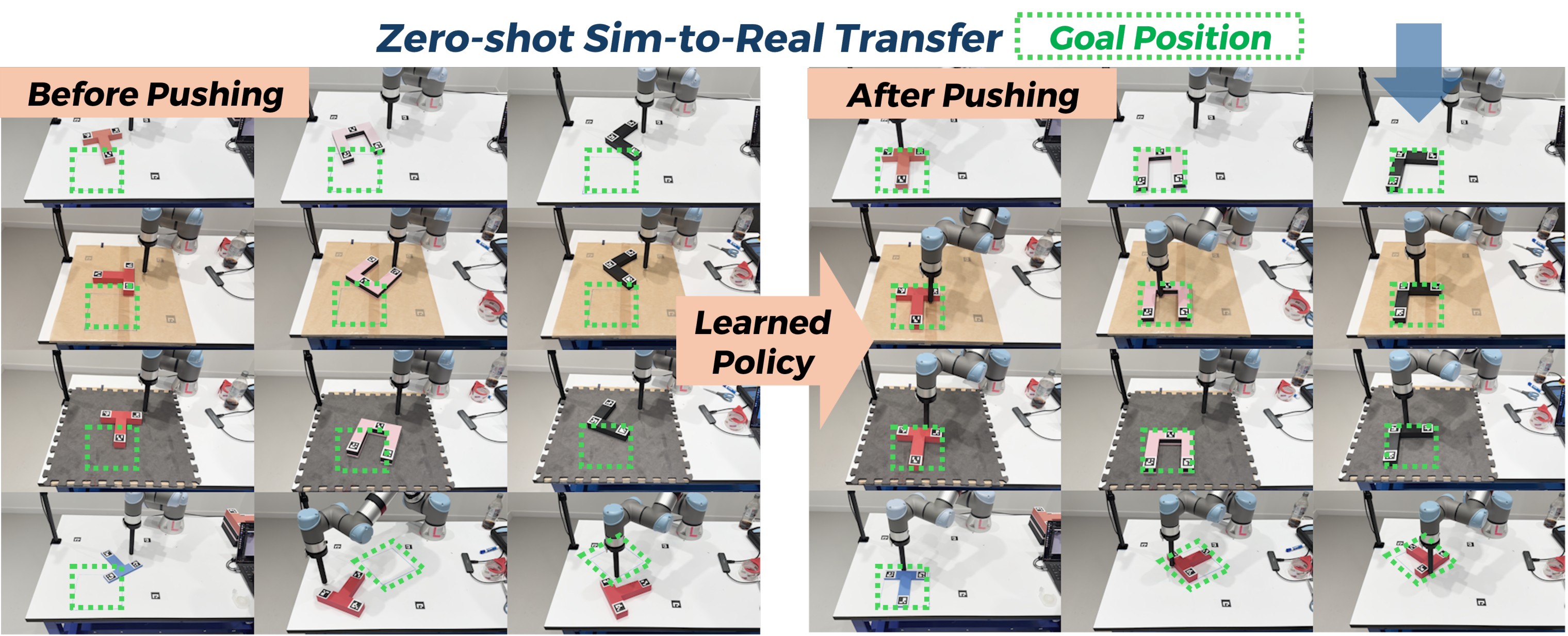}
    \caption{\small \textbf{(Upper)} The pipeline we propose to train the diffusion policies with RL to solve the multi-task block-pushing task. \textbf{(Lower)} The learned diffusion policy can achieve zero-shot transfer to real-world block-pushing tasks with different block shapes, block weights, frictions, and goal poses. More figures and videos are available at \url{https://diffusionpolicyrl.site/}.}
    \label{fig:placeholder}
\end{figure*}

\textbf{Contribution.} To tackle these challenges and study how diffusion policies can help to promote sim-to-real transfer of online RL, we adopt the widely-used block-pushing tasks as our benchmark and extend it to a multi-task setting by introducing additional objectives involving blocks of varying shapes and weights. Under this setting, we present a complete pipeline that trains diffusion policies to perform multi-task, contact-rich manipulations with zero-shot sim-to-real transfer using an online RL framework, which incorporates carefully crafted observation and action spaces, curriculum learning strategies and perception algorithms for better performance.
The contributions of this paper is summarized as follows:
\begin{itemize}[leftmargin=*,topsep=1pt,itemsep=0pt]
    \item[\textbf{a)}] We present a novel, simplified diffusion policy optimization method derived from policy mirror descent with a simple objective function, which only adds an additional reweighting factor to the evidence lower bound~\cite{ho2020denoising}. The reweighting factor is just the exponentiated action-value functions, which can seamlessly serve as the policy learning module in RL algorithms.
    
    \item[\textbf{b)}] We demonstrate that, by augmenting inputs with task information, employing objective-centric state representations, and leveraging reverse curriculum generation to mitigate exploration challenges, we can train a single diffusion policy to solve multiple block-pushing tasks involving diverse shapes and goal positions. In contrast, current RL methods with Gaussian policies (such as SAC and PPO) perform poorly even with these techniques.
    
    \item[\textbf{c)}] We conduct extensive evaluations of the learned diffusion policy on real-world block-pushing tasks across diverse setups. The learned policy achieves zero-shot transfer to the real-world setting under various weights, frictions and perturbations, showing adaptive and robust manipulation capabilities for these block-pushing tasks.
\end{itemize}
\section{Background}

\subsection{Multi-Task Block-Pushing}

Block-pushing is a representative contact-rich manipulation task, first introduced in~\cite{chi2023diffusion} as Push-T and gradually gaining popularity for benchmarking diffusion policies~\cite{hoeg2024streaming,li2024crossway}. The block-pushing task is challenging because \textbf{a)} the agent needs to learn how to make and break contact with the block to complete the task; \textbf{b)} the optimal solution is usually multi-modal; and \textbf{c)} the contact point is usually an unstable equilibrium that magnifies the sim-to-real gap.

To facilitate the study of multi-task learning performance, we extend Push-T to a multi-task setup that includes multiple block shapes (e.g., T-, C-, and L-shaped blocks) pushed to different goal positions, as illustrated in Figure \ref{fig:placeholder}.


\subsection{Reinforcement Learning} Reinforcement Learning (RL) is a standard framework for sequential decision making in Markov Decision Processes (MDPs). Formally, an MDP is described by a tuple $(\mathcal{S}, \mathcal{A}, P, r, \rho, \gamma)$, where $\mathcal{S}$ is the state space, $\mathcal{A}$ is the action space, $P(s' | s, a)$ is the transition probability function, $r(s, a)$ is the reward function, $\rho\in\Delta(\Scal)$ is the initial distribution, and $\gamma \in [0,1)$ is the discount factor. The objective of RL is to find a policy $\pi: \mathcal{S}\to \Delta(\mathcal{A})$ that maximizes the expected cumulative return $\mathbb{E}_\pi \left[ \sum_{t=0}^\infty \gamma^t r(s_t, a_t) \right]$.

\textbf{Policy Mirror Descent.} In this paper, we focus on policy optimization with a KL-divergence regularization term, also known as policy mirror descent~\cite{tomar2020mirror}, which leads to practical algorithms like trust region policy optimization (TRPO)~\cite{schulman2015trust}. Specifically, at each step, the policy is updated by solving the following optimization problem:
\begin{equation}
    \arg\max_{\pi} \mathbb E_{a\sim\pi}\bigl[ Q^{\pi^{\rm old}}(s, a) \bigr] -\lambda D_{\rm KL}(\pi\|\pi^{\rm old}), \label{eq:pmd}
\end{equation}
where $\pi^{\rm old}$ is the current policy, and $Q^{\pi^{\rm old}}(s,a):=\EE_{\pi^{\rm old}}$ $\bigl[ \sum_{t=0}^\infty\gamma^t r(s_t,a_t)|s_0=s,a_0=a \bigr]$ denotes its state-action value function.
Intuitively, the KL-divergence regularization guarantees that the updated policy should remain close to the current policy $\pi^{\rm old}$ after a single update. Solving \eqref{eq:pmd} yields a closed-form optimal solution
\begin{align}\label{eq:optimal_pi_mirror_descent}
    \pi(a|s) &= \pi^{\rm old}(a|s)\frac{\exp\bigl( Q^{\pi^{\rm old}}(s, a)/\lambda \bigr)}{Z(s)},\nonumber\\
    \text{where } Z(s) &= \int \exp\rbr{Q^{\pi^{\rm old}}(s, a)/\lambda} \pi^{\rm old}(a|s)da. 
\end{align}
Despite its seemingly simple form, the closed-form solution \eqref{eq:optimal_pi_mirror_descent} is usually intractable for fitting or sampling due to the appearance of the intractable normalization constant $Z(s)$. In fact, existing approaches that build upon parameterized policy families generally need to overcome this intractability issue by projecting the updated policy $\pi$ back into the policy family (via, e.g., minimizing $D_{\rm KL}(\pi||\pi_\theta)$), especially with the popular choice of Gaussian policy families~\cite{haarnoja2018soft,peng2019advantage,nair2020awac}. Such projection restricts the expressiveness of the policy family and hence hurts the performance of the learned policy, which can be avoided by diffusion policies.

\textbf{Difference from Behavior Cloning (BC) methods~\cite{chi2023diffusion}.} Existing BC-based diffusion policies generally learn to generate state or action trajectories implemented in a receding horizon manner. In contrast, when using RL-based diffusion policies, we only generate one action at each step. Meanwhile, instead of learning a visuomotor policy that directly predicts actions from images, we use state-based input and leverage a separate perception module to estimate the target pose. These techniques feature better interpretability, more efficient training, and more robust sim-to-real transfer. 

\subsection{Diffusion Policy}
A diffusion policy is, in essence, a conditional denoising diffusion probabilistic model (DDPM)~\cite{ho2020denoising} of action $a$ conditioned on the state $s$, aiming to represent the policy $\pi(a|s)$. DDPM consists of a manually designed forward process that gradually corrupts the original data distribution $a_0\sim\pi(\cdot|s)$ to Gaussian noise $a_T\sim\Ncal(0, I)$, and a reverse denoising process leveraging a noise prediction model $\boldsymbol{\epsilon}_\theta(s, a_t, t)$ that iteratively denoises the Gaussian noise $a_T$ to original data $a_0$, using the following denoising scheme
\begin{equation*}
  a_{t-1}=\frac{1}{\sqrt{\alpha_t}}\left(a_t-\frac{1-\alpha_t}{\sqrt{1-\bar{\alpha}_t}} \boldsymbol{\epsilon}_\theta\left(a_t, s, t\right)\right)+\sigma_t z_t
\end{equation*}
for $t=T,T-1,\dots, 1$. Here $z_t\sim\Ncal(0, I)$ and $\alpha_t,\bar\alpha_t,\sigma_t$ are constants depending on $t$ known as the noise schedule parameters.
To learn the noise prediction model $\boldsymbol{\epsilon}_\theta(a_t, s,t)$, \cite{ho2020denoising} shows that maximizing the evidence lower bound (ELBO) is equialent to minimizing the following loss function:
    \begin{equation}
    \resizebox{0.91\linewidth}{!}{\(
        \mathbb{E}_{\begin{subarray}{l} s\sim \Dcal, t\in\{1, 2,\dots, T\},\\a_0\sim \pi(\cdot|s),\epsilon\sim\Ncal(0, I) \end{subarray}}\sbr{\nbr{ \boldsymbol{\epsilon}_\theta\rbr{\sqrt{\bar\alpha_t}a_0 + \sqrt{1-\bar\alpha_t}\epsilon, s,  t} -  \epsilon}^2},
    \)}\label{eq:ddpm_loss}
    \end{equation}
where $\Dcal$ is the dataset (also known as replay buffer in RL).


\section{RL with Diffusion Policy}
In this section, we introduce the key components of the proposed multi-task RL method using diffusion policy, featuring a simple reweighted loss function, a unified design of observation and action spaces across different block-pushing tasks, and a curriculum learning scheme.

\subsection{Simplified Loss Function for Diffusion Policies}

In order to train a diffusion policy that fits the target policy defined in \eqref{eq:optimal_pi_mirror_descent} and overcome the aforementioned challenges for RL-based diffusion policy training, we design the energy-guided score matching (ESM) loss as follows:
\begin{equation}
    \resizebox{0.91\linewidth}{!}{\(
        \mathbb{E}_{\begin{subarray}{l} s\sim \Dcal, t\in\{1, 2,\dots, T\},\\ a\sim\pi^{\rm old},\epsilon\sim\Ncal(0, I) \end{subarray}}\left[\exp\left(Q^{\pi^{\rm old}}(s, a) / \lambda\right)\|\epsilon_\theta(a_t;s,t)-\epsilon\|^2\right],
    \)} \label{eq:weighted_loss}
\end{equation}
where $a_t = \sqrt{\bar\alpha_t}a_0 + \sqrt{1-\bar\alpha_t}\epsilon$ is consistent with \eqref{eq:ddpm_loss}, and the \emph{energy function} represents an importance sampling ratio as $Q^{\pi^{\rm old}}$ appears in the log-density ratio between $\pi$ and $\pi_{\rm old}$.

\begin{proposition}
The energy-guided score matching loss is equivalent to the evidence lower bound in \eqref{eq:ddpm_loss} that trains a diffusion policy to fit the target policy in~\eqref{eq:optimal_pi_mirror_descent}. 
\end{proposition}
The derivations are deferred to Appendix \ref{sec.appendix.derivations}. Note that the evidence lower bound in \eqref{eq:ddpm_loss} of target policy $\pi$ in~\eqref{eq:optimal_pi_mirror_descent} since we cannot directly sample from $\pi$ in~\eqref{eq:optimal_pi_mirror_descent}, thus the ESM loss is necessary for practical training of diffusion policies.

\textbf{Connections to existing diffusion-based RL methods.} The ESM loss function specified in~\eqref{eq:weighted_loss} is similar to the weighted score matching techniques used in several recent diffusion-based RL algorithms~\cite{hansen2023idql,zhang2025energy,ding2024diffusion}. As compared to IDQL~\cite{hansen2023idql} and QIPO~\cite{zhang2025energy}, our method avoids learning a baseline function for the weights. This is beneficial in that baseline estimation usually relies on Monte-Carlo sampling, which is biased and increases the computational cost. On the other hand, QVPO~\cite{ding2024diffusion} reweights the score matching by a truncated $Q$-function rather than the exponential $Q$-function, which fails to handle the negative values of $Q$-functions. In contrast, our ESM can effective handle these issues.

\subsection{Objective-centric State Representation for Observations and  Actions}\label{sec:obact}
In multi-task block-pushing tasks, it is important to carefully design the observation and action spaces to reduce the task complexity for a single universal policy. Here, we introduce our observation and action space designs by leveraging the objective-centric state representations.
The observation space is a combination of spatial states $s^{\rm states}\in\mathbb{R}^6$ and shape encoding $s^{\rm task}\in[0.0, 2.0]^4$, while the action space is a velocity command of the push agent $a\in[-1.0, 1.0]^2$.

\vspace{-3mm}
\begin{figure}[H]
\centering
    \includegraphics[width=0.4\linewidth]{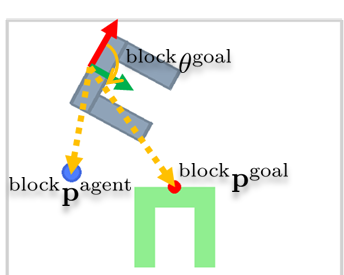}
    \includegraphics[width=0.4\linewidth]{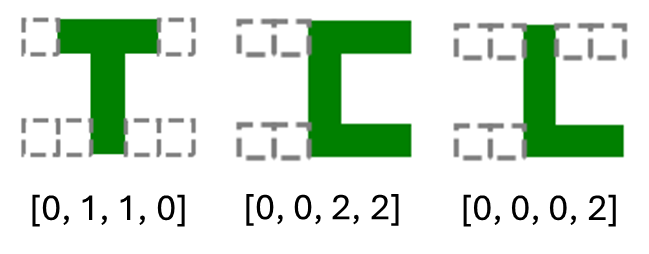}
  \vspace{-2mm}
  \caption{Observation space design: objective-centric representation \textbf{(left)} and shape encoding \textbf{(right)}.}
  \label{fig:obs}
\end{figure}
\vspace{-3mm}

\textbf{Implicit goal conditioning by objective-centric state representations.}
It is standard to condition the policy on the goal position when there are multiple goals to reach~\cite{liu2022goal}. However, simply conditioning on the absolute spatial position of the goal implicitly neglects the hidden prior knowledge of spatial transformation symmetry, making it hard for the learned policy to generalize to some simple unseen tasks (e.g., when the goal and initial positions undergo the same translational movement).
To design policies generalizable to arbitrary goal poses and improve data efficiency, we transform the observation and action spaces to the local coordinate of the block (``block frame''), without explicitly conditioning on the goal. The state vector is constructed as
$
s^{\rm state} = \{{}^B\mathbf{p}^A,{}^B\mathbf{p}^G, \sin({}^B\theta^G),\cos({}^B\theta^G)\}
$,
where ${}^B\mathbf{p}^A$ is the position of the agent, while ${}^B\mathbf{p}^G,{}^B\theta^G$ denote the position and orientation of the goal, \emph{both observed from the block frame at the current step}. Note that ${}^B\mathbf{p}^G,{}^B\theta^G$ are updated at every step, so the task information is implicitly embedded in these observations in local coordinates. 

\textbf{Shape Encoding.} In addition, we also condition the policy on a shape-encoding vector $s^{\rm shape}\in[0.0, 2.0]^4$, as shown in Figure~\ref{fig:obs}. Note that the naive shape encoding used here is just a proof-of-concept, which can be easily replaced by advanced 2D or image-based shape encoding. Experimental results in Section~\ref{sec.exp_results} show that, even with such a simple scheme, diffusion policies are already able to display promising zero-shot transfer capabilities to unseen shapes.

\textbf{Action space.} The action is the velocity observed in the goal frame ${}^G\mathbf{v}^A$, which can transformed to world frame by $\mathbf{v}^A = {}^WR^G{}^G\mathbf{v}^A$. Here $ {}^WR^G$ is the rotation matrix of the goal pose with respect to the world frame. 

\subsection{Curriculum Learning via Reverse Curriculum Generation}\label{sec:sampling}

\begin{figure}[H]
  \centering
  \includegraphics[width=0.7\linewidth]{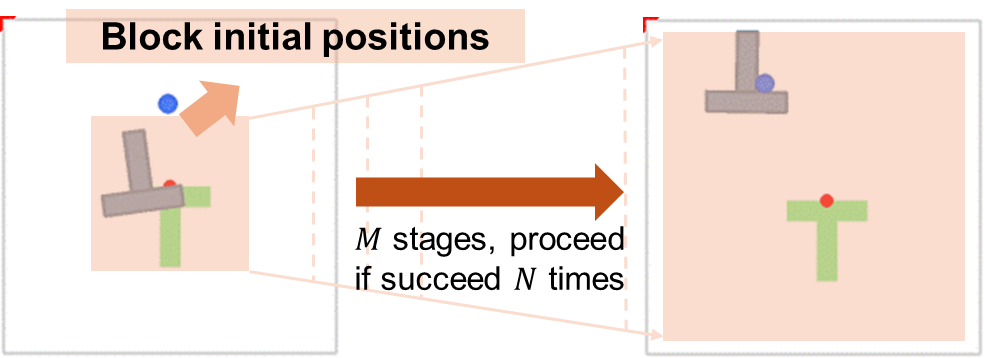}
  \vspace{-2mm}
  \caption{Curriculum learning by expanding the distance to goal.}
  \label{fig:curriculum}
\end{figure}

We follow an idea similar to the reverse curriculum generation~\cite{florensa2017reverse}, where the curriculum is designed to gradually increase the distance from the goal position to the initial position, as shown in Figure \ref{fig:curriculum}. The curriculum is divided into $M$ stages, and is updated automatically when the push succeeds for $N$ times. More implementation details of the curriculum learning scheme can be found in Appendix \ref{sec.appendix.implementations}.

\vspace{-2mm}
\vspace{-2mm}

\textbf{Comments on the sample efficiency improvement techniques.} Curriculum learning and spatial information design require minimal domain knowledge of the tasks, which should be generalizable to a variety of goal-reaching manipulation tasks. Our ablation studies in Section \ref{sec.ablations} show that they are vital to the RL performance  for block-pushing.
\section{Real-world Implementation}

\subsection{Zero-shot Sim-to-Real Transfer}

\textbf{Benefits of our framework design.} The state-based inputs required by our algorithm allow easy sim-to-real transfer as long as we have access to an accurate pose estimation algorithm, which in practice can be implemented using ArUco markers or point cloud registration algorithms. 
Moreover, the observation and action space design facilitates training with a fixed goal pose in the simulator, while still enabling generalization to multiple different goal positions in the real world, owing to the relative nature of the state vector $s^{\rm state}$. 

\textbf{Significance of sim-to-real gaps.} In our setup, the major sim-to-real gap of the block-pushing task lies in the friction coefficients against the surface. However, real-world experimental results (see Section~\ref{sec.real_exp}) confirm that the learned diffusion policy demonstrates consistently high performance in pushing the block across three distinct surface types. It can be argued that, with a powerful policy, the sim-to-real gap will not pose a significant challenge to the performance as long as it does not affect system stability, in which case (including our block-pushing tasks) we can simply ignore the sim-to-real gap and achieve zero-shot sim-to-real transfer.

\subsection{Block Pose Estimation}
Unlike other diffusion policies~\cite{chi2023diffusion}, our diffusion policy leverages state-based inputs. Therefore, we need to estimate the pose of the blocks before feeding the state into the policy, for which we shall resort to markers or vision-based methods. 

\textbf{ArUco markers.} In this setup, ArUco markers are affixed to the blocks to provide robust pose estimation for robot manipulation, as shown in Figure~\ref{fig:real_setup} below. 

\textbf{Point cloud registration.} Alternatively, we also develop a vision-based point cloud registration algorithm to avoid the usage of ArUco markers. Unlike many other objects in manipulation, our blocks have flat surfaces and sharp edges, which is incompatible with the commonly practice of normal-vector-based registration methods. Alternatively, we propose to use a geometry-based method that leverages the regular shape of the blocks to project them to the table surface before performing 2D point cloud registration. 

\vspace{-3mm}
\begin{figure}[H]
    \centering
    \includegraphics[height=0.298\linewidth]{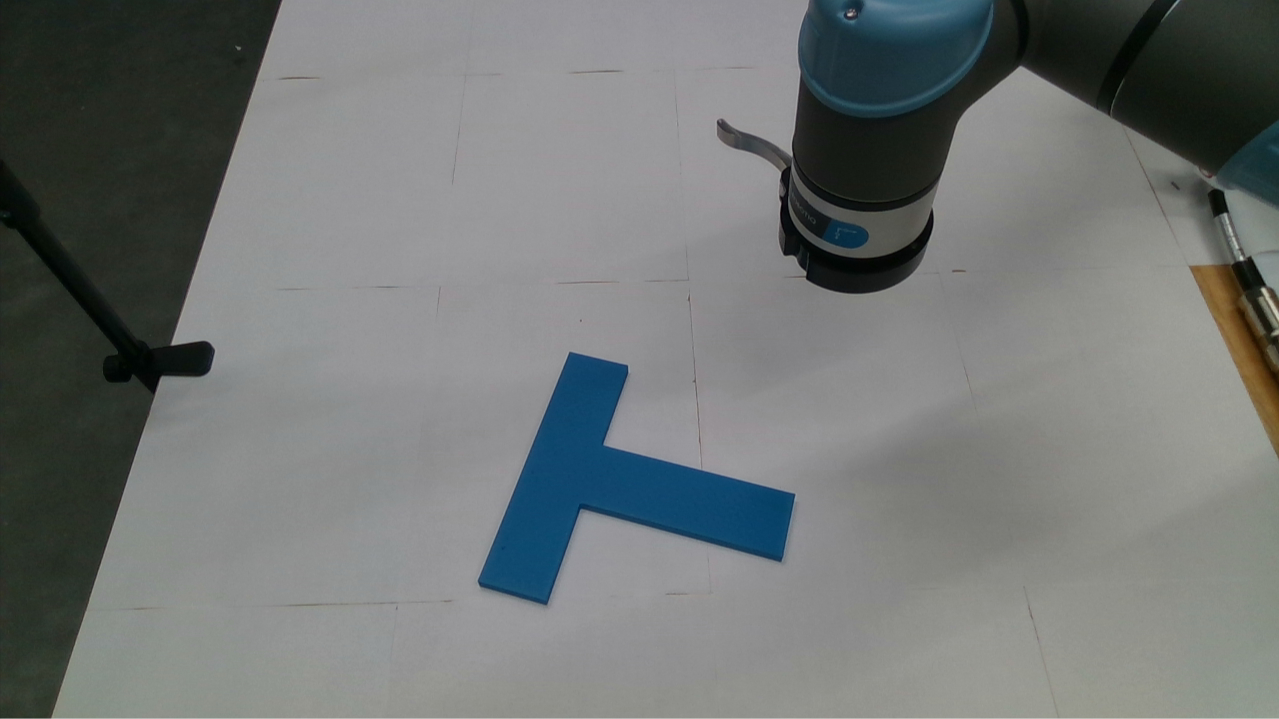}\includegraphics[height=0.3\linewidth]{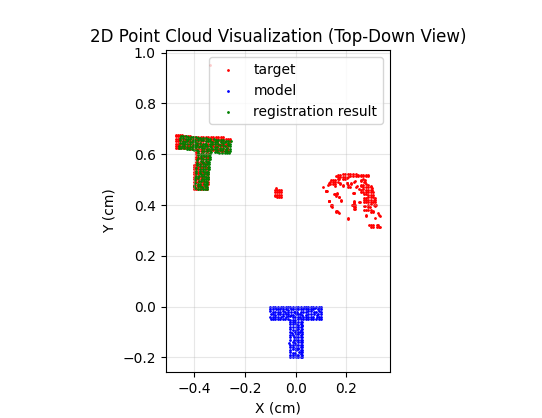}\\
    \vspace{-1mm}
    \caption{An example point cloud registration results when large additional noises exist in the segmentation results. The point cloud transformed by the estimated transformation~\textbf{(green)} fits well with the true point cloud projected to the table plane~\textbf{(red)}.}
    \label{fig:point_cloud}
\end{figure}
\vspace{-3mm}

\textbf{Additional filters and simulators.} The primary challenges to accurate pose estimation arise from observation noise and occlusions caused by the robot arm during motion. To handle the additional observation noise, we add a Kalman filter on top of the pose estimation results. Moreover, in the case when blocks are significantly blocked by the robot arm, we run a simulator parallel to the pose estimation module as a backup option, and switch to simulated pose estimations when the occlusion obstructs accurate pose estimation.
\section{Experimental Results}

In this section, we present a comprehensive experimental evaluation of our algorithm. We first show an extensive comparison in the simulated environment, and then proceed to the setup and results of real-world implementation.

\subsection{Simulation Results}

\textbf{Simulation setup.} 
We use the same simulator as in~\cite{chi2023diffusion,florence2022implicit} based on PyMunk physics simulators. We adapt the simulator to match our observation and action space definitions described in Section~\ref{sec:obact}. 
The algorithm is implemented using the JAX framework. All environments are trained with a total of 4 million environment interactions under 3 random seeds. After training, we evaluate each algorithm in 100 randomized test trials to evaluate the algorithm by success rate and episodic lengths (algorithm terminates upon success; smaller episodic length indicates better performance).

\subsubsection{Simulation Experiment Results}\label{sec.exp_results}\quad

\textbf{Simulation performance and baselines.} We compare the proposed ESM algorithm against three families of baselines. \textbf{a)} Classic model-free RL algorithms: PPO~\cite{schulman2017proximal}, TD3~\cite{fujimoto2018addressing}, and SAC~\cite{haarnoja2018soft}, where PPO and SAC use Gaussian policy and TD3 uses deterministic policy. \textbf{b)} Recent
diffusion-based RL algorithms: QVPO~\cite{ding2024diffusion}, QSM~\cite{psenka2023learning}, and DACER~\cite{wang2024diffusion}. 
\textbf{c)} Behavior cloning~\cite{chi2023diffusion}: we directly use the model checkpoints released by the authors. Since simulation data is abundant and cheap to obtain, we do not consider offline RL as baselines. 
The results are shown in Table \ref{tab:results}, which shows that only our ESM algorithm and SAC successfully learn policies that can solve the block-pushing task. Further, ESM achieves 100\% block-pushing success rate and much lower episode lengths compared to SAC as well as all other baselines.

\vspace{-2mm}
\begin{table}[h]
\centering
\caption{Performance comparison of methods on the multi-shape pushing task (maximum episode length: 300).}
\label{tab:results}
\vspace{-6pt}
\resizebox{\linewidth}{!}{
\begin{tabular}{llcc}
\toprule
 & Method & \textbf{Success Rate (\%)~$\uparrow$}  & \textbf{Episode Length~$\downarrow$}     \\
\midrule
\multirow{3}{*}{\begin{tabular}{@{}l@{}}Classic\\Model-free\\ RL\end{tabular}} 
    & SAC~\cite{haarnoja2018soft} & $ 66.3 \pm 12.2 $ & $118.9 \pm 130.0 $   \\
    & TD3~\cite{fujimoto2018addressing} & $ 0\pm 0$ & $ 300 \pm 0$ \\
    & PPO~\cite{schulman2017proximal} & $ 0\pm 0$ & $ 300 \pm 0$   \\
\midrule
\multirow{3}{*}{\begin{tabular}{@{}l@{}}Diffusion\\Policy RL\end{tabular}} 
    & ESM (ours) & $ \mathbf{100 \pm 0} $ & $\mathbf{21.1 \pm 5.6}$\\
    & QVPO~\cite{ding2024diffusion} & $ 0 \pm 0$ & $300 \pm 0$\\
    & DACER~\cite{wang2024diffusion} & $ 0 \pm 0$ & $300 \pm 0$\\
    & QSM~\cite{psenka2023learning} & $ 0 \pm 0$ & $300 \pm 0$  \\
    \midrule
 \makecell{Imitation \\ Learning} & Diff. Policy~\cite{chi2023diffusion}$^*$  & $ 35$ & $179.54 \pm 34.12$\\
\bottomrule
\end{tabular}
}
\end{table}
\vspace{-2mm}

\begin{wrapfigure}{r}{0.08\textwidth}
  \vspace{-10pt}
  \includegraphics[width=1.0\linewidth]{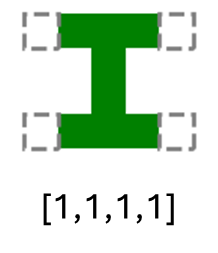}
  \vspace{-20pt}
  \caption{Unseen I-shape to test generalization.}
  \label{fig:ishape}
  \vspace{-10pt}
\end{wrapfigure}

\begin{figure}[t]
    \centering
    \includegraphics[width=0.49\linewidth]{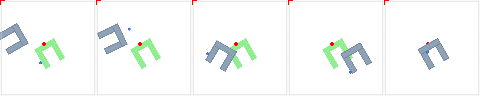}
    \includegraphics[width=0.49\linewidth]{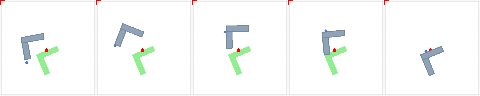}
    \includegraphics[width=0.49\linewidth]{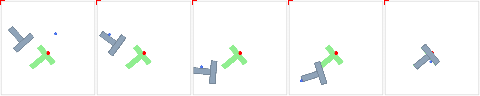}
    \includegraphics[width=0.49\linewidth]{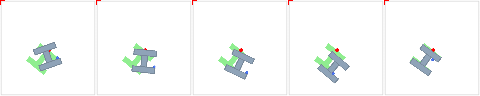}
    
    \caption{Screenshots of zero-shot generalization of the proposed ESM policy to random goal poses and unseen I-shape (last one located lower right).}
    \label{fig:screenshots}
    \vspace{-5mm}
\end{figure}

\textbf{Multi-shape pushing and generalization to unseen shape and goals.}
We evaluate our method separately on different blocks to further justify its multi-task capabilities. To assess generalization beyond the training distribution, we further evaluate the methods on an unseen I-shaped block without any additional training. Our ESM is compared against the best baseline in previous expriments, SAC, to fully display its advantage. The results are reported in Table~\ref{tab:multi_shape_results}.
It is observed that ESM achieves consistently high performance ($100\%$) on the training shapes, and moderately high performance ($61\%$) on the unseen I-block, demonstrating strong generalization without any additional training. On the contrary, the performance of SAC displays high variance, with particularly poor performance (complete failure) on the T-block and the unseen I-block. Screenshots of the block-pushing behavior are shown in Figure~\ref{fig:screenshots}, highlighting the robustness and superior generalization ability of ESM compared to SAC.

\vspace{-2mm}
\begin{table}[h]
\centering
\caption{Success rates of SAC and ESM on different shapes.}
\label{tab:multi_shape_results}
\vspace{-6pt}
\begin{tabular}{lcccc}
\toprule
\textbf{Method} & \textbf{T-block} & \textbf{C-block} & \textbf{L-block} & \textbf{I-block (zero-shot)}  \\
\midrule
SAC   & 0\% & 100\% & 99 \% & 0\% \\
ESM  & 100\% & 100\% & 100\% & 61\%  \\
\bottomrule
\end{tabular}
\end{table}

\subsubsection{Ablation Studies}\label{sec.ablations}

\begin{figure}[h]
    \centering
    \vspace{-3mm}
    \includegraphics[width=0.49\linewidth]{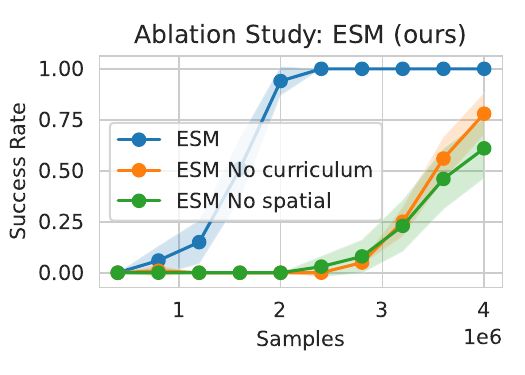}
    \includegraphics[width=0.49\linewidth]{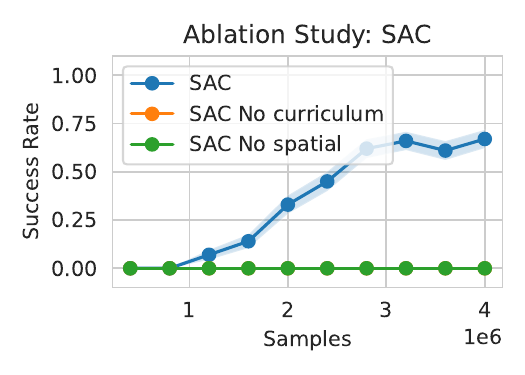}
    \vspace{-7mm}
    \caption{Ablation studies of ESM (proposed) and SAC (best baseline). The proposed ESM consistently perform better on multiple ablations.}
    \label{fig:ablations}
    \vspace{-4mm}
\end{figure}

We conduct ablation studies to justify the necessity and contribution of the efficiency improvement techniques specified in Section \ref{sec:sampling}, including:

\begin{itemize}[leftmargin=*,topsep=1pt,itemsep=0pt]
    \item \textbf{Remove curriculum learning.} Under this ablation, the block is randomly initialized in the whole space.

    \item \textbf{Remove objective-centric state representations.} Under this ablation, the observation consists of agent, block and goal poses, all observed from the \emph{world frame}.
\end{itemize}
Both ablation studies compare the performance of the proposed ESM against the best baseline SAC, the result of which are shown in Figure \ref{fig:ablations}. It can be seen that the proposed techniques effectively improve the data efficiency and performance of the diffusion policy training. Moreover, even after we ablate the curriculum learning or spatial observation design, diffusion policies can still achieve acceptable success rates (as compared the the complete failure of SAC), showing a huge performance improvement of the proposed ESM compared to the SAC baseline.

\subsection{Real-world Experiments}\label{sec.real_exp}

\begin{wrapfigure}{r}{0.24\textwidth}
  \centering
  \vspace{-4mm}
  \includegraphics[width=0.99\linewidth]{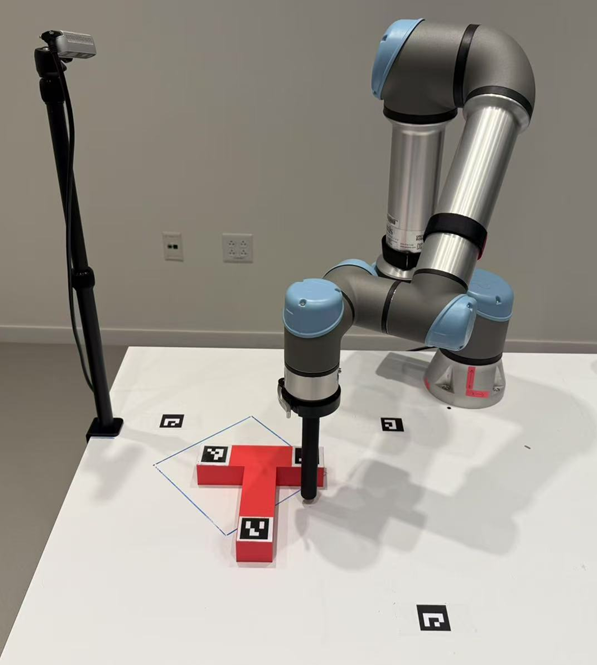}
  \caption{Real-world configuration of the block-pushing tasks.}
  \label{fig:real_setup}
  \vspace{-1mm}
\end{wrapfigure}
\textbf{Setup.} We evaluate the real-world performance of the learned diffusion policy using the setup shown in Figure \ref{fig:real_setup}. Specifically, we use a UR5e manipulator to push the blocks and an Intel RealSense D435  camera placed upper left to localize the block. The light blue square drawn on the table indicates the goal position. We use the original 3D printing models for the stick and the T-block from~\cite{chi2023diffusion}, and create our own models of other shapes.

\begin{figure}[h]
    \centering
    \includegraphics[width=\linewidth]{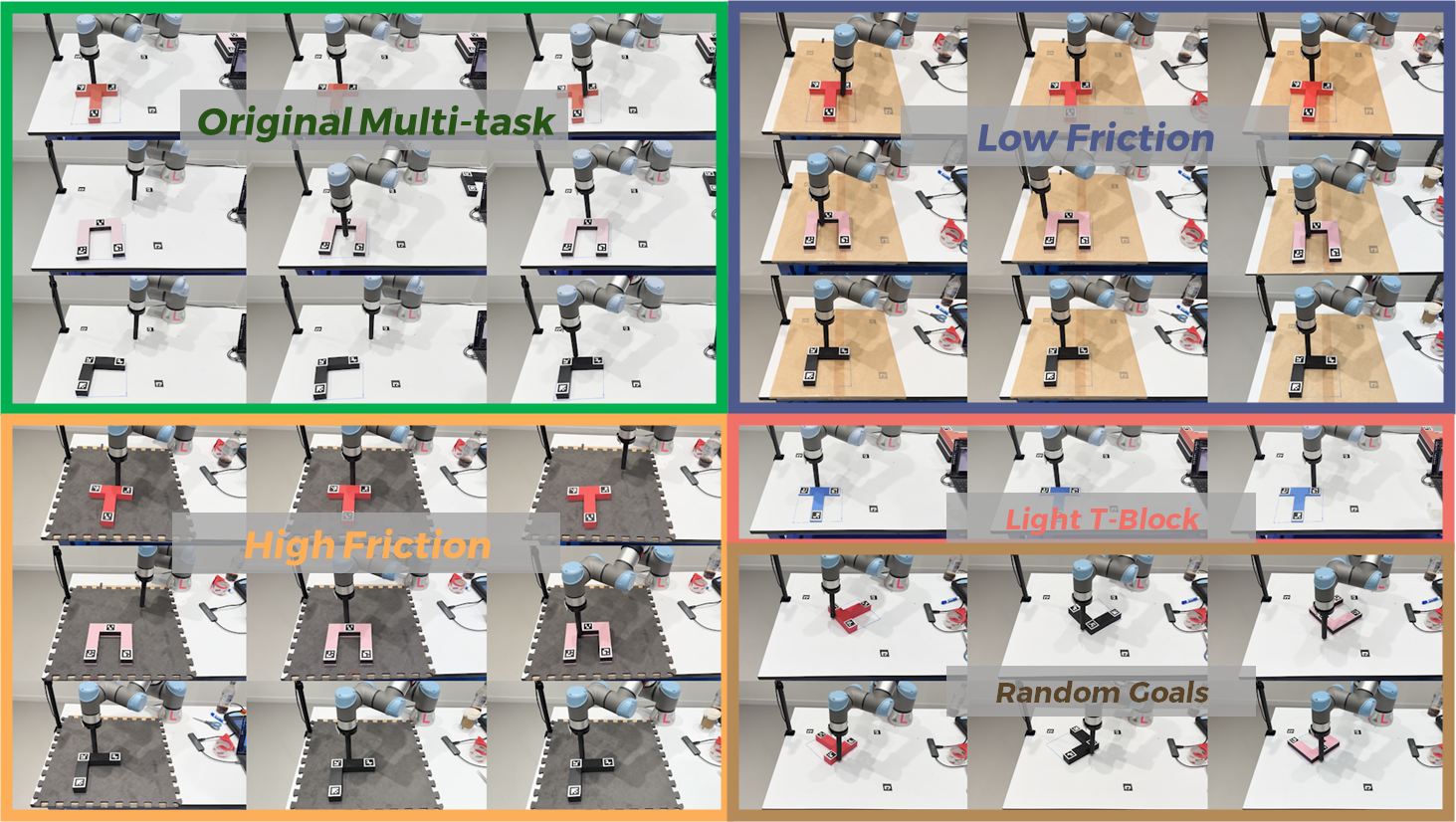}
    \caption{All evaluation tasks in the real-world setting. Figures show the final timestep with successful pushing. Some initial states are shown in Figure~\ref{fig:placeholder}. 
    }
    \label{fig:real_all}
\end{figure}

\textbf{Tasks.} As shown in Figure \ref{fig:real_all}, the learned diffusion policy is evaluated in the real-world setting for a total of 36 tasks under varying conditions: \textbf{a) Friction.} To test the robustness against different friction levels, we use floor mats for higher friction levels and non-stick parchment paper for lower friction levels. \textbf{b) Block weight.} We print another thinner T-block that is only a quarter as thick (1 cm thick; the blue T-block in Figure~\ref{fig:real_all}) as the original T-block (4 cm; see the red block). \textbf{c) Goal poses.} We also test the policy using two random goal poses different from the one used in training.

\vspace{-2mm}
\vspace{-3mm}

\textbf{Zero-shot transfer performance.} We record the zero-shot transfer performance of the learned diffusion policy, which can be found at our website~\url{https://diffusionpolicyrl.site/}. The learned diffusion policy \emph{successfully solves all the tasks}, showing its powerful capability for sim-to-real transfer and strong robustness to real-world conditions.
Moreover, we also test the baseline policy learned by SAC on the 9 training tasks (shown in the upper left in Figure \ref{fig:real_all}). A comparison of the success rate, total distance traveled, and total steps to push the block to goal position are listed in Table \ref{tab:real_world}. It is evident that the proposed ESM algorithm has a significant advantage over SAC with a Gaussian policy in terms of their real-world performances.

\vspace{-2mm}
\begin{table}[H]
\caption{Comparison of ESM and SAC in 9 real-world tasks.}\label{tab:real_world}
\centering
\vspace{-2mm}
\begin{tabular}{@{}ccccc@{}}
\toprule
Shape & Metric & Success & Distance (m) & Steps \\ \midrule
\multirow{2}{*}{T} 
 & ESM & \textbf{3/3} & 1.0  & 15.67 \\
 & SAC & 0/3$^*$ & - & - \\
\midrule
\multirow{2}{*}{C} 
 & ESM & \textbf{3/3} & \textbf{1.11} & \textbf{17.67} \\
 & SAC & \textbf{3/3} & 2.48 & 40.33 \\
\midrule
\multirow{2}{*}{L} 
 & ESM & \textbf{3/3} & \textbf{1.39} & \textbf{22.67} \\
 & SAC & \textbf{3/3} & 1.46 & 24 \\
\bottomrule
\end{tabular}\\
{\footnotesize $\quad ^*$ SAC failed in pushing T, so the distance and steps are not counted.}
\end{table}
\vspace{-3mm}

\textbf{Empirical observations.} After a detailed examination and comparison of the zero-shot sim-to-real transfer behavior produced by the proposed ESM and the baseline SAC algorithms, we observe that: \textbf{a)} in the simulated setting, the agent is able to learn multiple skills (such as pushing and rotating) using both algorithms, and can further use a diverse combination of these skills to finish the tasks; \textbf{b)} in the real-world setting, as some moves induce larger sim-to-real gaps, the agent falls back to a specific sequence of movement to finish the tasks. This empirical observation prompts us to further consider how the multi-modality of diffusion policies benefits the sim-to-real performance of RL algorithms. 

\section{Related works}

\textbf{Diffusion policy.} In the realm of robotics manipulation, diffusion policies~\cite{ho2020denoising,sohl-dickstein2015deep,song2021scorebased} have been widely leveraged to generate long-horizon robot trajectories, either using behavior cloning training schemes~\cite{janner2022planning,chi2023diffusion,du2024learning} or mimicking expert policies in MDPs using offline RL~\cite{wang2022diffusion,chen2022offline,hansen2023idql,zhang2025energy}, both requiring an abundance of expert demonstrations. The expressiveness of diffusion policies also brings high multi-task capabilities~\cite{ni2023metadiffuser,he2023diffusion,du2024learning}, but a lack of expert data becomes the major bottleneck to scale up these multi-task diffusion policies. 
Recent findings have enabled diffusion policy in online RL~\cite{ding2024diffusion,psenka2023learning,wang2024diffusion,zhang2025energy}, yet they still suffer from various issues including biased solution approximation and expensive multi-step gradient preparations, which undermines the true potential of diffusion policies in online RL.

\textbf{Goal-conditioned RL and curriculum learning.} Goal-conditioned RL refers to such RL approaches that aim to reach or manipulate the agent to reach a set of goal states that are usually not unique~\cite{liu2022goal}, which is challenging due to its multi-task and sparse-reward nature. To solve these issues, existing methods rely on reward shaping with estimated distance metrics~\cite{andrychowicz2017hindsight} or success probability estimation~\cite{eysenbach2022contrastive,chane2021goal}. However, these techniques usually require complex algorithm design or involving domain knowledge. Curriculum learning~\cite{bengio2009curriculum} is another promising alternative that requires less domain knowledge, which has proven to be helpful in solving challenging RL tasks including goal reaching and drone stabilization~\cite {narvekar2018learning,narvekar2020curriculum,eschmann2024learning}.


\section{Conclusion}
In this paper, we deploy a novel RL framework to demonstrate that one single diffusion policy trained with online RL can successfully handle complex, multi-task, contact-rich block-pushing tasks without reliance on expert demonstration data. By imposing an energy-based value function guidance on the training loss, incorporating curriculum learning, and carefully designing observation and action spaces, our approach effectively learns high-performance diffusion policies from sparse rewards for challenging tasks where traditional RL with unimodal Gaussian policies tend to fail. Empirical evaluations further reveal diffusion policies' strong zero-shot transfer capabilities to real-world tasks, with diverse variations in goal positions, block shapes, block weights and friction levels, highlighting the robustness and efficiency of RL with diffusion policy for challenging manipulation tasks.

\textbf{Limitations.} The zero-shot sim-to-real transfer is sensitive to the sim-to-real gaps. Although our results show that the sim-to-real gap is not significant for block-pushing on 2D surfaces, it could be significant for other manipulation tasks. Potential approaches to close the sim-to-real gap include domain randomization with zero-shot transfer~\cite{tobin2017domain} and residual learning for dynamics and action models for few-shot transfer and continual learning~\cite{saveriano2017data,he2025asap}. The proposed RL pipeline can also be used in fine-tuning current models built upon BC or other imitation learning schemes. 







\bibliographystyle{IEEEtran}
\bibliography{example}
\appendix
\renewcommand{\ab}{a}
\renewcommand{\sbb}{s}
\renewcommand{\Ib}{I}
\subsection{Derivations of Energy-Guided Score Matching}\label{sec.appendix.derivations}

We first revisit the closed-form optimal solution of the DSM loss~\eqref{eq:ddpm_loss}, and then show that our ESM loss~\eqref{eq:weighted_loss} induces the same optimal solution to conclude loss equivalence.

\subsubsection{Noise prediction to learn noise-perturbed score function} We first revisit an existing result showing that the optimal solution of the noise prediction model $\epsb_\theta(a_t,s,t)$ is $-\sigma_t\nabla_{\ab_t}\log \tilde\pi_t(\ab_t|\sbb)$, i.e. the \emph{noise-perturbed score function}~\cite{shribak2024diffusion}. Recall that the term \emph{score function} refers to $\nabla_x \log p(x)$, the gradient of the log-density function.

Before stating the result, we need to first define the noise-perturbed policy. For any fixed $\sbb$, consider action samples $\ab_0 \sim \pi(\cdot|\sbb)$ corrupted by a Gaussian kernel $q_{t|0}(\ab_t|\ab_0)$ $ =\mathcal{N}(\ab_t;\sqrt{\bar\alpha_t}\ab_0, \rbr{1 - \bar\alpha_t}\Ib)$ under any noise schedule $\{ \bar\alpha_t \}$, resulting in the noise-perturbed policy $\tilde\pi_t(\cdot|\sbb)$ with density
\begin{equation*}
    \tilde\pi_t(\ab_t|\sbb)= \int q_{t|0}(\ab_t|\ab_0)\pi(\ab_0|\sbb)d\ab_0,\quad t=1,2,\dots,T.
\end{equation*}
Note that the Gaussian corruption kernel implies that $a_t = \sqrt{\bar\alpha_t}a_0 + \sqrt{1 - \bar\alpha_t}\epsilon$, where $\epsilon\sim\mathcal{N}(0, I)$. This leads to the connection between the noise $\epsilon$ and the score function, i.e.,
\begin{align*}
\nabla_{a_t}\log q_{t|0}(\ab_t|\ab_0)  = -\frac{\epsilon}{\sqrt{1-\bar\alpha_t}}
\end{align*}
For convenience, we use the shorthand $\sigma_t:=\sqrt{1-\bar\alpha_t}$.

\begin{proposition}[Diffusion models learn noise-perturbed score functions\label{prop:diff_ebm}, \cite{vincent2011connection}]
By optimizing the DSM loss in \eqref{eq:ddpm_loss}, 
the score network ${ \epsilon_\theta(\ab_t;\sbb, t)}$ matches ${-\sigma_t\nabla_{\ab_t}\log \tilde\pi_t(\ab_t|\sbb)}$ for all ${a_t}$ in the action space ${\mathcal{A}}$,
which is the noise-perturbed score functions scaled by factor $-\sigma_t$.
     \end{proposition}

\subsubsection{ESM loss as an equivalent reweighted loss.}
Consider an equivalent loss with reweighting function $g$, i.e.,
\begin{equation}
    \displaystyle \Lcal^g(\theta, s, t) = \int g(a_t,s,t)\nbr{\epsb_\theta(\ab_t;\sbb, t) +\sigma_t\nabla_{\ab_t}\log \tilde \pi_t(\ab_t|\sbb_t)}^2da_t\label{eq:general_g_loss},
\end{equation}
where $g(a_t,s,t)$ is a strictly positive function everywhere in the action space $\mathcal{A}$ for any fixed $s$ and $t$. The loss is called ``equivalent'' since we can indeed show that any reweighted loss will result in the same optimal solution as the one for the original DSM loss specified in Proposition~\ref{prop:diff_ebm}. Specifically, when we choose 
$g(a_t,s,t) = \tilde\pi(a|s)Z(s)$, we have
\begin{align*}
        &\Lcal^g(\theta, s, t) \\\
        {}={}& \int g(a_t,s,t)\nbr{\epsb_\theta(\ab_t;\sbb, t) +\sigma_t\nabla_{\ab_t}\log \tilde \pi_t(\ab_t|\sbb_t)}^2da_t\\
        & (\textit{Plug in the proof of Proposition \ref{prop:diff_ebm} in \cite{vincent2011connection}})\\
        {}={}& \int g(a_t,s,t)\frac{1}{\tilde\pi_t(\ab_t|\sbb)}\int q_{t|0}(\ab_t|\ab_0)\pi(\ab_0|\sbb) \cdot{} \\
        &\quad\nbr{\epsb_\theta(\ab_t,\sbb, t) -\epsilon}^2  d\ab_0da_t  + \texttt{const}\\
        & (\textit{Apply the definition of }\pi \text{ in \eqref{eq:optimal_pi_mirror_descent}})\\
        {}={}& \int g(a_t,s,t)\frac{1}{\tilde\pi_t(\ab_t|\sbb)}\int q_{t|0}(\ab_t|\ab_0)\pi^{\rm old}(a_0|s) \cdot{} \\
        &\quad\frac{\exp\bigl( Q^{\pi^{\rm old}}(s, a_0)/\lambda \bigr)}{Z(s)} \nbr{\epsb_\theta(\ab_t,\sbb, t) -\epsilon}^2  d\ab_0da_t  + \texttt{const}\\
        &(\textit{Apply the definition of }g),\\
        {}={}& \iint q_{t|0}(\ab_t|\ab_0)\pi^{\rm old}(a_0|s)\exp\bigl( Q^{\pi^{\rm old}}(s, a_0)/\lambda \bigr) \cdot{} \\
        &\quad \nbr{\epsb_\theta(\ab_t,\sbb, t) -\epsilon}^2  d\ab_0da_t  + \texttt{const}\\
        {}={}& \EE_{\begin{subarray}{k} a_0\sim\pi^{\rm old},\\ \epsilon\sim\mathcal{N}(0, I) \end{subarray}}\sbr{\exp\bigl( Q^{\pi^{\rm old}}(s, a_0)/\lambda \bigr)\nbr{\epsb_\theta(\ab_t,\sbb, t) -\epsilon}^2 } +\texttt{const}.
\end{align*}
Integrate the above over $s$ and $t$, and we recover exactly the ESM loss specified in~\eqref{eq:weighted_loss}. This completes the proof.

\subsection{Algorithm Implementation and Experimental Details}\label{sec.appendix.implementations}

\textbf{Detailed design of curriculum learning.} At the $m$\textsuperscript{th} stage, the initial distribution is designed such that $x, y$ are randomly sampled from $ [256 - (30 + m/M \cdot 100), 256 + (30 + m/M \cdot 100)]$, with $(256, 256)$ defined as the center of the canvas. We select $M=10,N=50$ in our implementation.




\textbf{Comparison with Behavior Cloning (BC).}
We compare ESM against behavior cloning (BC) based on expert demonstrations using the simulator and experimental protocol specified in~\cite{chi2023diffusion}. We train ESM in the same simulator environment  using the proposed RL objective. To evaluate~\cite{chi2023diffusion}, we use the model checkpoints released by the authors. Then we test each method across $100$ randomized scenarios. Table~\ref{tab:method_comparison} reports the average episode lengths and success rates achieved by both methods. It is observed that ESM significantly outperforms the BC baseline, and also exhibits superior performance compared to SAC, both in terms of higher success rate and shorter episode length on average. Notably, ESM successfully completes the task in all $100$ test scenarios without a single failure.

\vspace{-2mm}
\begin{table}[h]
\centering
\caption{Performance comparison in the simulator provided by~\cite{chi2023diffusion}.}
\label{tab:method_comparison}
\vspace{-3mm}
\begin{tabular}{lcc}
\toprule
\textbf{Method}  & \textbf{Success Rate (\%) $\uparrow$}& \textbf{Episode Length $\downarrow$} \\
\midrule
BC~\cite{chi2023diffusion}  & $ 35$ & $179.54 \pm 34.12$\\
SAC~\cite{haarnoja2018soft}& $87$  &$ 64.49 \pm 59.44$ \\
ESM (ours) & $\mathbf{100}$ &$\mathbf{24.61\pm 8.62}$ \\
\bottomrule
\end{tabular}
\end{table}

\end{document}